\pdfoutput=1

\documentclass[11pt]{article}

\usepackage[preprint]{acl}

\usepackage{times}
\usepackage{latexsym}
\usepackage{amsmath}
\usepackage{amssymb}
\usepackage{multirow}
\usepackage{tabularx}
\usepackage{listings}
\usepackage{multirow}
\usepackage{booktabs}
\usepackage{url}
\usepackage{enumitem}
\usepackage{placeins}

\usepackage[T1]{fontenc}

\usepackage[utf8]{inputenc}

\usepackage{microtype}

\usepackage{inconsolata}

\usepackage{graphicx}

\newcommand{\aaron}[1]{{\textcolor{orange}{\bf [{\sc aaron:} #1]}}}

\title{GLiREL - Generalist Model for Zero-Shot Relation Extraction}

\author{
  Jack Boylan, Chris Hokamp, Demian Gholipour Ghalandari \\
  Quantexa \\
  \texttt{<firstname><lastname>@quantexa.com} \\}

\begin{document}
\maketitle
\begin{abstract}
We introduce GLiREL (\textbf{G}eneralist \textbf{L}ightweight model for zero-shot \textbf{Rel}ation Extraction), an efficient architecture and training paradigm for zero-shot relation classification. Inspired by recent advancements in zero-shot named entity recognition, this work presents an approach to efficiently and accurately predict zero-shot relationship labels between multiple entities in a single forward pass. Experiments using the FewRel and WikiZSL benchmarks demonstrate that our approach achieves state-of-the-art results on the zero-shot relation classification task. In addition, we contribute a protocol for synthetically-generating datasets with diverse relation labels.
\end{abstract}

\section{Introduction}
Recent advances in zero-shot NLP models for entity recognition have been enabled by large-scale synthetic training data generation using state-of-the-art (SoTA) Large Language Models (LLMs) \cite{zhou2024universalnertargeteddistillationlarge-pile-ner}. An ongoing line of work achieves drastic improvements in accuracy and usability over previous approaches by using efficient architectures targeted at various NLP tasks \citep{bogdanov2024nunerentityrecognitionencoder, stepanov2024glinermultitaskgeneralistlightweight, zaratiana2023gliner}. Zero-shot named entity recognition (NER) models such as GLiNER \citep{zaratiana2023gliner} do not operate on a fixed label set, only requiring textual labels to be specified at inference time, and can directly perform span classification using labels that are not observed during training. 


\begin{figure}[htp]
  \centering
  \resizebox{\columnwidth}{!}{%
    \includegraphics{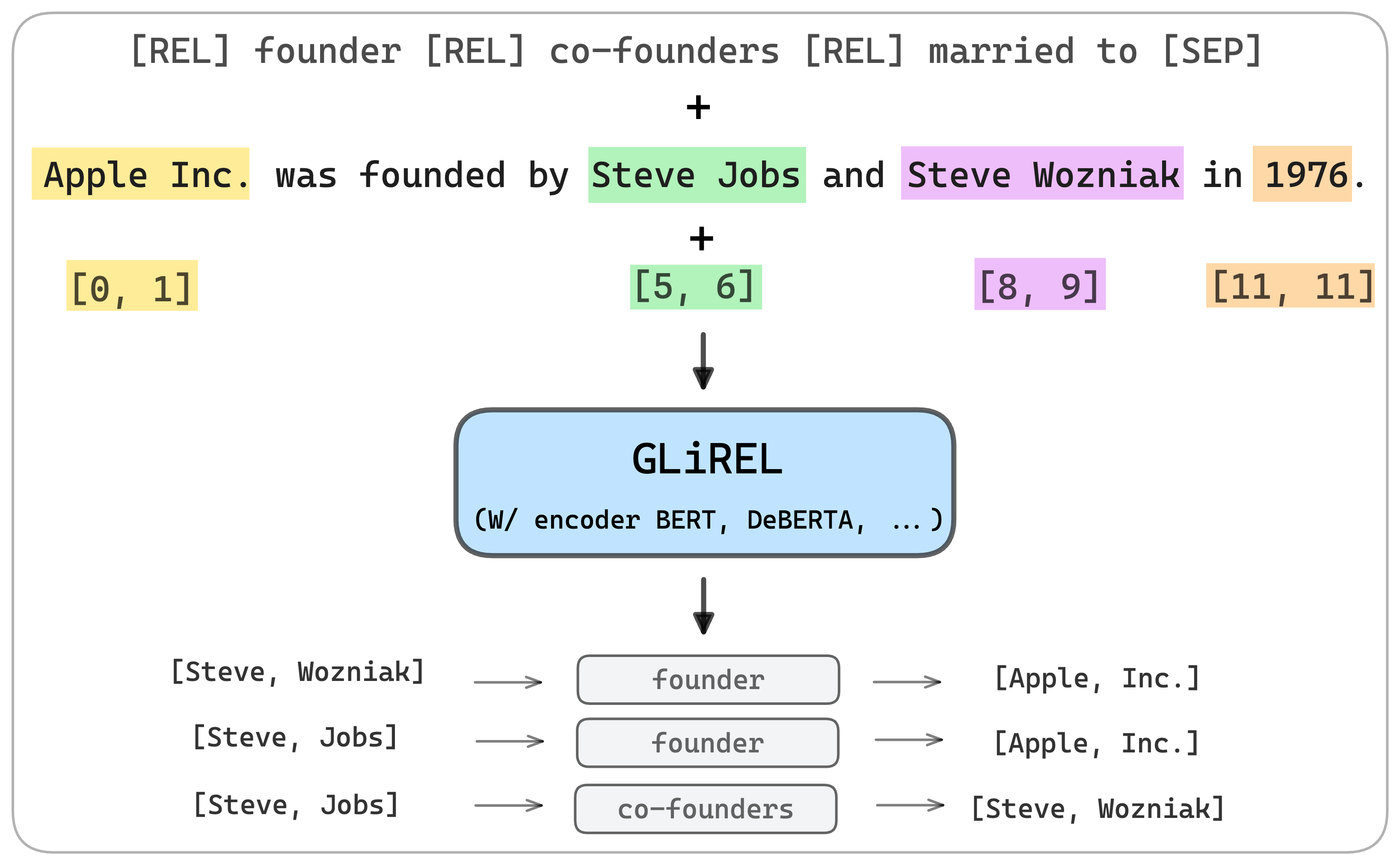}
  }
  \caption{Example inputs and outputs for GLiREL.}
  \label{fig:shiny-input}
\end{figure}

In contrast to generative models, targeted architectures for zero-shot span classification jointly predict all labels simultaneously, making them much more efficient than auto-regressive models \cite{zaratiana2023gliner}. Existing SoTA zero-shot relation classification\footnote{The terms \textit{relation classification} and \textit{relation extraction} are used interchangeably throughout the literature.} (ZSRC) models achieve strong performance, but are inefficient because every entity pair and candidate label combination is treated as a separate input. Existing methods do not scale to real-world use cases, where a large number of entity pairs is  extracted from text, each of which must be classified against many candidate labels. GLiREL takes inspiration from recent successes in zero-shot NER and text classification, adapting these approaches to enable ZSRC that is both efficient and accurate.

While SoTA LLMs excel at information extraction (IE) tasks \cite{li2024metaincontextlearningmakes, zhou2024universalnertargeteddistillationlarge-pile-ner}, there are major limitations to their scale and deployment patterns, including:
\begin{itemize}
\item Auto-regressive decoding is unable to take advantage of task-specific parallelism,
\item Specific, expensive hardware requirements,
\item Output is not sufficiently constrained unless guided by heuristic decoding methods,
\item Unpredictable behavior, for example when asked to identify \textit{all} relationships between entities in a document of arbitrary length.
\end{itemize}

\begin{figure*}[htp]
  \centering
  \resizebox{\textwidth}{!}{%
    \includegraphics{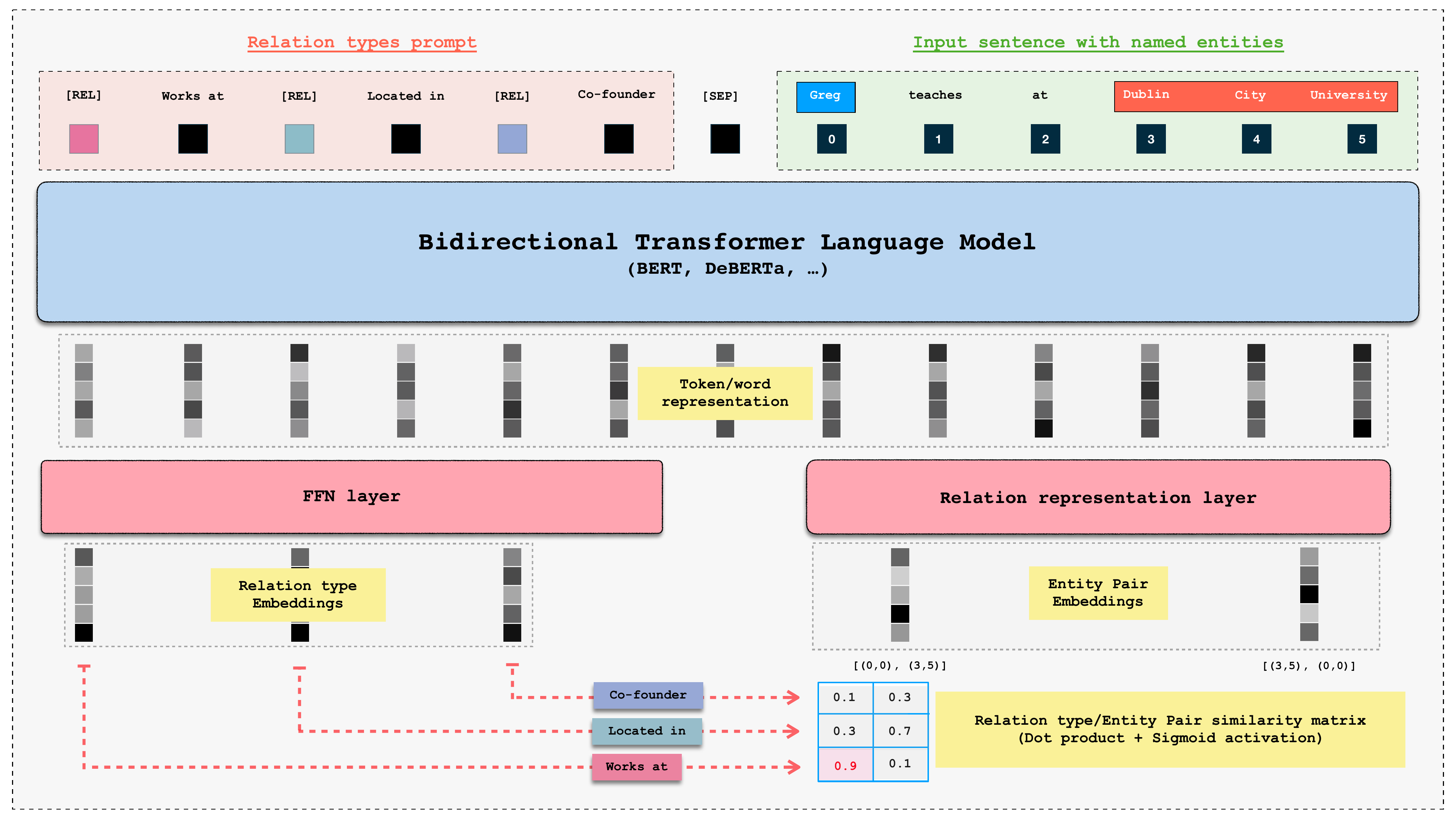}
  }
  \caption{Our proposed approach to zero-shot relation extraction. Firstly, the relation labels and N entities are encoded using a bidirectional transformer. The N entity embeddings will be concatenated to form $N^2$ pairs in the relation representation layer. The relation labels are fed through a feed-forward network to obtain relation type embeddings. A scoring layer then computes the similarity between every label and every entity pair. Diagram adapted from \citet{zaratiana2023gliner}.}
  \label{fig:main-diagram}
\end{figure*}

\noindent The ability of LLMs to perform zero-shot inference with unconstrained output makes them very flexible, but for many tasks, their auto-regressive factorization raises issues around reliability. For information extraction tasks in particular, assigning relationships between all entities found in a text is a requirement that cannot be achieved reliably with LLMs without auxiliary models \cite{li2024llmrelationclassifierdocumentlevel} and/or at-runtime data augmentation mechanisms \cite{jiang-etal-2024-relation, Ma_2023}. 

In contrast to the disadvantages above, LLMs excel at \textit{unconstrained} classification and labeling tasks, and can be effectively utilized to produce large scale datasets for training downstream models with task-specific inductive biases. Synthetic dataset generation using general-purpose LLMs is a critical component of the recent success of zero-shot NLP models \citep{zaratiana2023gliner,bogdanov2024nunerentityrecognitionencoder}. This paper includes a protocol for generating a large-scale dataset for relationship classification which enables training zero-shot models. The key contributions of our work are:

\begin{itemize}[itemsep=0.3pt, topsep=3pt]
    \item A novel zero-shot relation extraction architecture, GLiREL.
    \item A training dataset construction policy that results in a high-quality synthetic dataset for training zero-shot relation classification models.
    \item A training paradigm producing SoTA results on zero-shot relation extraction benchmarks.
\end{itemize}

\noindent The paper is organized as follows: section \ref{sec:background} discusses important related work, section \ref{sec:method} discusses the GLiREL model architecture, section \ref{sec:experiments} presents our experimental results, and sections \ref{sec:ablations}, \ref{sec:conclusion} and \ref{sec:limitations} provide discussion, analysis, and examinations of the limitations of this work. All code is publicly available.\footnote{\url{https://github.com/jackboyla/GLiREL}}

\section{Background}
\label{sec:background}

\paragraph{Joint vs Independent NER and Relation Classification}

Joint entity and relationship classification \cite{Eberts2019SpanbasedJE,zaratiana2024enrico} can enhance performance through task transfer and global optimization, but increases decoding complexity and reduces flexibility, often requiring bespoke architectures that may not generalize to other tasks. In contrast, traditional IE pipelines use independent models (e.g., spaCy \cite{Honnibal_spaCy_Industrial-strength_Natural_2020}), offering flexibility but making relationship classification dependent on static upstream NER components. Our work assumes entities are provided by an upstream component and focuses on detecting relationships between these entities using zero-shot relation labels, thus maintaining pipeline flexibility and allowing classification between any number and types of entities from diverse sources.





\paragraph{Zero-Shot Relation Extraction} 
\label{bkg:zs-re}

Zero-shot relation extraction is an appealing avenue of research because of the flexibility and simplicity of the inference and training paradigms. MC-BERT \cite{MC-BERT-Lan-et-al} can use previously-unseen relation type labels to classify entity pairs by treating the task as a multiple-choice problem. TMC-BERT \cite{moeller2024zeroshot} extends upon this by incorporating entity types and relation label descriptions. MC-BERT and TMC-BERT construct a template for each entity pair for each candidate label in an instance. This results in a large number of inputs from a relatively small sample size, making it unsuitable for scaling. RelationPrompt \cite{chia-etal-2022-relationprompt} generates synthetic training examples at inference time using GPT2, requiring a large number ($N=250$) of examples per label, which is resource-intensive. DSP uses a discriminative prompting strategy to classify both entities and relations in a zero-shot setting. ZS-SKA \cite{gong2024promptbased} performs ZSRC by using templates to augment data and incorporating an external knowledge graph. ZSRE \cite{tran2023enhancing} encodes text and relation labels separately, computing semantic correlation for each entity pair and label combination, leading to inefficiency in real-world scenarios where many entities are present.

In contrast, GLiREL supports any relation labels at inference time without lengthy descriptions or entity type information (see Figure \ref{fig:shiny-input}). Multiple entity pairs can be classified in a single input, making our approach more efficient than models that require multiple rounds of inference or at-runtime data generation. Additionally, GLiREL processes relation labels and input text simultaneously, capturing interactions between all labels and entity pairs.


\paragraph{LLMs for Relation Classification}

LLMs have been leveraged for relation classification, achieving strong zero- and few-shot performance using meta in-context learning and synthetic data generation \cite{li2024metaincontextlearningmakes,xu2023unleash}. Some approaches reformulate zero-shot relation classification as a question-answering task \cite{li2023revisitinglargelanguagemodels}. In document-level RE (DocRE), finetuning LLaMA2 with LoRA shows significant improvements, especially when a pretrained language model first classifies whether an entity pair expresses a relationship before passing it to the LLM \cite{li2024llmrelationclassifierdocumentlevel}. GenRDK \cite{sun2024consistencyguidedknowledgeretrieval} uses chain-of-retrieval prompts with ChatGPT to generate synthetic data for finetuning LLaMA2. Alternatively, \citet{xue2024autoredocumentlevelrelationextraction} finetune an LLM to propose head and tail entities given a document and relation label, outperforming other LLM-based baselines.

\paragraph{Zero-Shot Learning and Synthetic Training Data Generation}
\citet{zaratiana2023gliner} and \citet{bogdanov2024nunerentityrecognitionencoder} showed that a straightforward and efficient model architecture can achieve excellent performance on the zero-shot NER task, given high-quality, large scale training data. Open-source LLMs have enabled the creation of this kind of training data, through simple and scalable protocols which prompt a model to label the entities in a short text with any type label \cite{zhou2024universalnertargeteddistillationlarge-pile-ner}. Importantly, labels are not constrained to a particular taxonomy, and the generative model is free to assign any representative label to entities in the text. In the case of GLiNER \cite{zaratiana2023gliner}, training on the Pile-NER dataset (created by \citet{zhou2024universalnertargeteddistillationlarge-pile-ner}) enabled a new SoTA in zero-shot NER. 

\section{Method}
\label{sec:method}

The GLiREL architecture has three main components: 

\begin{itemize}[itemsep=0.5pt, topsep=2pt]
    \item A pre-trained bidirectional langage model used as the \textbf{text encoder}, which jointly processes candidate relations and input texts.
    \item An \textbf{entity pair representation module} which extracts vector representations for all entities in the text and proposes a representation for every pair of entities.
    \item A \textbf{scorer module} to compute the similarity between entity pair representations and relation label representations.
\end{itemize}

\noindent The architecture encodes relation labels and entity pair embeddings in the same latent space to compute their similarity. The overall architecture is illustrated in Figure \ref{fig:main-diagram}. We choose DeBERTa V3-large as the encoder model due to its excellent performance on downstream tasks \cite{he2023debertav3improvingdebertausing}.

\subsection{Input}
\label{sec:input}

The model input sequence is comprised of an ordered list of elements, where an element is a string containing one or more tokens. Concretely, inputs are built from: 

\begin{itemize}[itemsep=0.5pt, topsep=5pt]
    \item a list of $M$ zero-shot labels denoted as $t_m$, each separated by a special \verb|[REL]| token: $t_0,$ \verb|[REL]|$, t_1,$ \verb|[REL]|$, ..., t_{M-1}$. Both $t_m$ and \verb|[REL]| are treated as elements,
    \item a special \verb|[SEP]| token to indicate the end of the labels prompt. \verb|[SEP]| is treated as a single element,
    \item the input text, denoted as a list of N tokens $x_0, x_1, ..., x_{N-1}$, where each token is an element.
\end{itemize}

\begin{figure}[htbp]
  \centering
  \resizebox{\columnwidth}{!}{
    \includegraphics{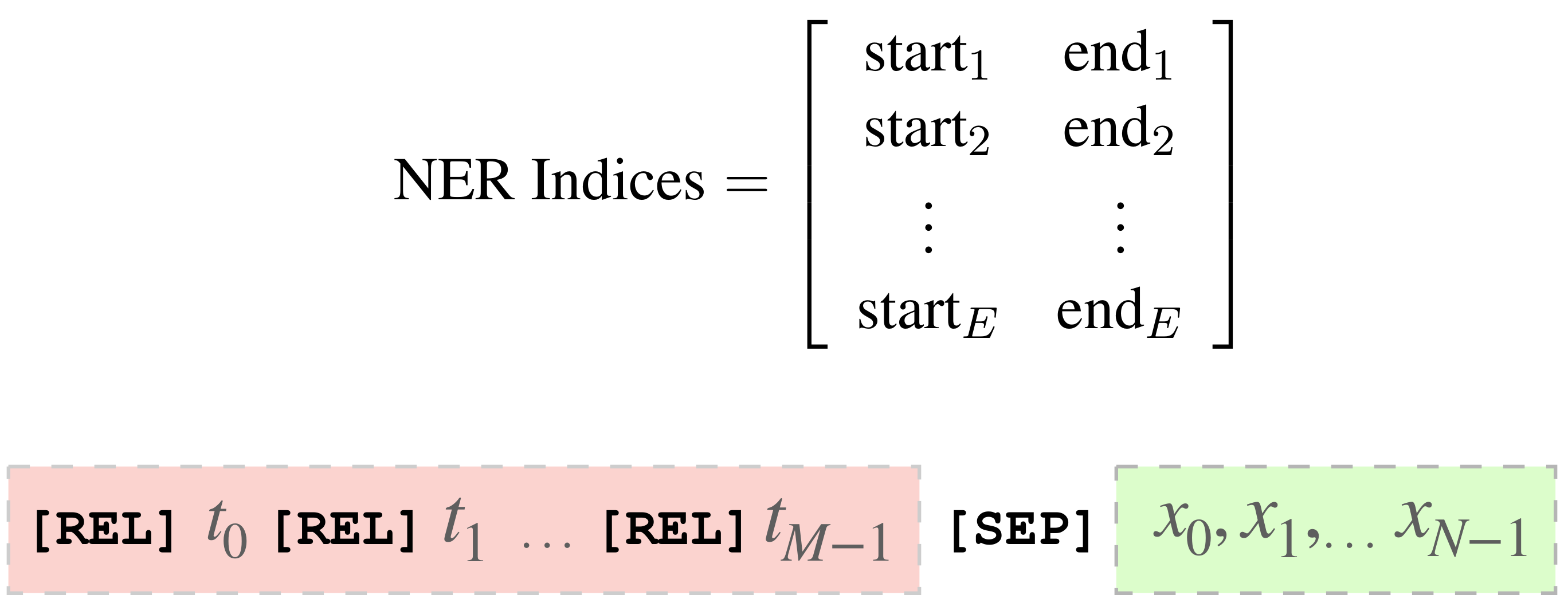}
  }
  \caption{GLiREL input includes relation types $t_{0}, ..., t_{M-1}$, text tokens $x_0, ..., x_{N-1}$, and the start and end indices of all entities within the text.}
  \label{fig:example-input}
  \vspace{-0.2cm}
\end{figure}

\noindent The GLiREL model additionally expects indices for $E$ known entities in the text, represented as pairs of start and end positions. 
The input structure is illustrated in Figure \ref{fig:example-input}.

\paragraph{Tokenization} 

The special \verb|[REL]| and \verb|[SEP]| tokens are added to the encoder's tokenizer vocabulary. The input sequence from Figure \ref{fig:example-input} is tokenized accordingly, ensuring that relation type labels and special tokens are properly handled. For this study, we follow the pooling strategy described in \citet{zaratiana-etal-2022-named} by taking the first subtoken representation of each element. Details of the tokenization process are provided in Appendix~\ref{appendix:tokenization}.

\subsection{Token Representation}

The token encoder processes the input sequence to compute interactions between all tokens (from both the relationship labels and from the input text), producing contextualized representations. Let \( \mathbf{p} = \{ \mathbf{p}_t \}_{0}^{M-1} \in \mathbb{R}^{M \times D} \) represent the encoder’s output for each relation type, corresponding to the first subtoken representation of each relation type label. Similarly, \( \mathbf{h} = \{ \mathbf{h}_i \}_{0}^{N-1} \in \mathbb{R}^{N \times D} \) denotes the representation of each word in the input text. As already mentioned, for words tokenized into multiple subwords we use the representation of the first sub-word.

\subsection{Label and Entity Pair Representation}
\label{sec:ent-pair-rep}

We aim to encode relationship labels and entity pair embeddings into a unified latent space. We follow the methodology of GLiNER \cite{zaratiana2023gliner}, with additional steps for entity pair representation and refinement layers.

\paragraph{Relation Label Representation:} 

After pooling, each relationship label in the input sequence is represented by a vector \( \mathbf{p}_i \), Relation label representations are additionally transformed by a two-layer feed-forward network (FFN) as shown in equation \ref{eq:relation-label-transformation}:

\vspace{-0.6cm}

\begin{equation}
\mathbf{q} = \text{FFN}(\mathbf{p}) = \{ \mathbf{q}_t \}_{t=0}^{M-1} \in \mathbb{R}^{M \times D},
\label{eq:relation-label-transformation}
\end{equation}

\noindent where \( M \) is the total number of relationship labels, and \( D \) is the dimensionality of the model's hidden layers. \( \mathbf{q}_t \) thus represents the transformed vector for the \( t^{th} \) relationship label.
    
\paragraph{Entity Representation:}

The entity indices given as input to the model (see Figure \ref{fig:example-input}) are used to extract entity representations from the word representations \( \mathbf{h} \). The representation of an entity starting at position \( i \) and ending at position \( j \) in the input text, \( \mathbf{e}_{ij} \in \mathbb{R}^{D} \), is computed as 

\begin{equation}
\mathbf{e}_{ij} = \text{FFN}(\mathbf{h}_i \otimes \mathbf{h}_j).
\label{eq:span-rep}
\end{equation}

\noindent In equation \ref{eq:span-rep}, FFN denotes a two-layer feed-forward network, and \( \otimes \) represents the concatenation operation.

\paragraph{Entity Pair Representation:}

Let \( \mathbf{e}_{u} = \mathbf{e}_{ij} \) represent the \( u^{th} \) entity representation computed in Equation \ref{eq:span-rep} using its start and end positions $i$ and $j$. For any distinct entity pair \( (u, v) \), where \( u \neq v \), the pair representation \( \boldsymbol{\kappa}_{uv} \) is computed as:

\begin{equation}
\boldsymbol{\kappa}_{uv} = \text{FFN}(\mathbf{e}_{u} \otimes \mathbf{e}_{v}), \quad \forall u \neq v
\end{equation}

\noindent where \( \otimes \) denotes the concatenation operation, and self-pairs are explicitly excluded. The concatenated entity pair representations are passed through a FFN for projection into the model’s latent space. The resulting representations \( \boldsymbol{\kappa}_{uv} \in \mathbb{R}^{D} \) are either further refined (Section \ref{sec:refine}), or used directly for scoring (Section \ref{sec:scoring-layer}). 

\begin{table*}[h]
\centering
\begin{tabular}{c|l|ccc|ccc|ccc}
\toprule
\multirow{2}{*}{$m$} & \multirow{2}{*}{Model} & \multicolumn{3}{c|}{Wiki-ZSL} & \multicolumn{3}{c|}{FewRel} \\
                     &                        & P     & R     & F1    & P     & R     & F1    \\
\midrule
\multirow{7}{*}{5} 
& RelationPrompt \cite{chia-etal-2022-relationprompt}         & 70.66 & 83.75 & 76.63 & 90.15 & 88.50 & 89.30 \\
& DSP-ZRSC \cite{lv-etal-2023-dsp}               & 94.10  & 77.10  & 84.80  & 93.40  & 92.50  & 92.90  \\
& ZSRE \cite{tran2023enhancing}    & \textbf{94.50} & \textbf{96.48} & \textbf{95.46} & 96.36 & \textbf{96.68} & \textbf{96.51} \\
& MC-BERT \cite{MC-BERT-Lan-et-al}                 & 80.28 & 84.03 & 82.11 & 90.82 & 91.30 & 90.47 \\
& TMC-BERT  \cite{moeller2024zeroshot}             & 90.11 & 87.89 & 88.92 & 93.94 & 93.30 & 93.62 \\
& GPT-4o$^\dagger$& 91.24& 72.07& 80.03& 96.75& 83.05&89.20\\
& GLiREL$^\dagger$& 69.88& 65.82& 62.80& 94.56& 89.17&81.21\\
& GLiREL (+ synthetic pretraining)$^\dagger$& 89.41& 80.67& 83.28& \textbf{96.84}& 93.41&94.20\\
\midrule
\multirow{7}{*}{10} 
& RelationPrompt \cite{chia-etal-2022-relationprompt}         & 68.51 & 74.76 & 71.50 & 80.33 & 79.62 & 79.96 \\
& DSP-ZRSC \cite{lv-etal-2023-dsp}               & 80.00  & 74.00  & 76.90  & 80.70  & 88.00  & 84.20  \\
& ZSRE \cite{tran2023enhancing}     & 85.43 & \textbf{88.14} & \textbf{86.74} & 81.13 & 82.24 & 81.68 \\
& MC-BERT \cite{MC-BERT-Lan-et-al}                & 72.81 & 73.96 & 73.38 & 86.57 & 85.27 & 85.92 \\
& TMC-BERT \cite{moeller2024zeroshot}              & 81.21 & 81.27 & 81.23 & 84.42 & 84.99 & 85.68 \\
& GPT-4o$^\dagger$& 77.62& 66.14& 68.35& 84.07& 58.00&66.20\\
& GLiREL$^\dagger$& 76.45& 71.80& 68.89& 85.40& 78.29&80.14\\
& GLiREL (+ synthetic pretraining)$^\dagger$& \textbf{89.87} & 81.56& 83.67& \textbf{91.09}& \textbf{87.42}&\textbf{87.60}\\
\midrule
\multirow{7}{*}{15} 
& RelationPrompt NG \cite{chia-etal-2022-relationprompt}      & 54.45 & 29.43 & 37.45 & 66.49 & 40.05 & 49.38 \\
& DSP-ZRSC  \cite{lv-etal-2023-dsp}              & 77.50  & 64.40  & 70.40  & 82.90  & 78.10  & 80.40  \\
& ZSRE \cite{tran2023enhancing}    & 64.68 & 65.01 & 65.30 & 66.44 & 69.29 & 67.82 \\
& MC-BERT  \cite{MC-BERT-Lan-et-al}              & 65.71 & 67.11 & 66.40 & 80.71 & 79.84 & 80.27 \\
& TMC-BERT \cite{moeller2024zeroshot}               & 73.62 & 74.07 & 73.77 & 82.11 & 79.93 & 81.00 \\
& GPT-4o$^\dagger$& \textbf{81.04}& 32.06& 41.57& 84.42& 65.76&70.70\\
& GLiREL$^\dagger$& 66.14& 65.40& 60.91& 75.76& 71.34&70.40\\
& GLiREL (+ synthetic pretraining)$^\dagger$& 79.44& \textbf{74.81}& \textbf{73.91}& \textbf{88.14}& \textbf{84.69}&\textbf{84.48}\\
\bottomrule
\end{tabular}
\caption{Performance comparison of models on Wiki-ZSL and FewRel datasets for various values of unseen relations \( m \). Metrics are averaged at the macro level. Values in \textbf{bold} are the best metrics for the given dataset and value of $m$. The dagger ($\dagger$) denotes our reported results; the remaining results are copied from their original papers. An extended table comparing more models can be found in appendix Table \ref{tab:full-results-appendix}.}
\label{tab:zs-relation-classification-results}
\end{table*}

\begin{table}[ht] 
\vspace{-0.1cm} 
\centering 
\begin{tabularx}{\columnwidth}{@{}X@{\hskip 5pt}c@{\hskip 5pt}c@{\hskip 5pt}c@{}} 
\hline
\textbf{Dataset} & \textbf{\# Instances} & \textbf{\# Rel Types} & \textbf{\# Triples} \\ 
\hline
Wiki-ZSL & 94,383 & 113 & 183,269 \\ 
FewRel & 56,000 & 80 & 56,000 \\ 
Re-DocRED & 4,053 & 96 & 120,664 \\
\hline
\end{tabularx} 
\caption{Dataset statistics.}
\label{tab:dataset-stats} 
\vspace{-0.5cm} 
\end{table}

\subsection{Refinement Layer}
\label{sec:refine}

The refinement layer is used to further process both the relation type representations \( \mathbf{q} \) and the entity pair representations \( \boldsymbol{\kappa}_{uv} \). Inspired by the filter and refine module in joint entity and relation extraction work from \citet{zaratiana2024enrico}, the refinement layer can be applied to:

\begin{itemize}[itemsep=0.5pt, topsep=5pt]
    \item Refine the entity pair representation with respect to the text,
    \item Refine the relation label representations with respect to the entity pairs, or
    \item Refine both.
\end{itemize}

\noindent The refinement process is composed of two main stages: (1) a cross-attention mechanism and (2) a feed-forward network (FFN) applied iteratively for a number of layers.

Given the entity pair representations \( \boldsymbol{\kappa}_{uv} \) and the relation type representations \( \mathbf{q}_t \), the refinement process can be written as follows:

\paragraph{Cross-Attention:} The representations are refined using cross-attention, where the entity pair representations attend to the relation type representations, and vice versa. For the entity pair refinement, we compute:

\begin{equation}
\boldsymbol{\kappa}_{uv}' = \boldsymbol{\kappa}_{uv} + \text{CrossAtt}(\boldsymbol{\kappa}_{uv}, \mathbf{q}_t)
\end{equation}

\noindent where \( \text{CrossAtt}(a, b) \) represents the cross-attention (also called encoder-decoder attention) mechanism as used by \citet{vaswani2023attentionneed}, which allows information exchange between \( a \) and \( b \).

\paragraph{Self-Attention:} 

The refined entity pair representation undergoes further refinement using a self-attention mechanism to capture intra-pair interactions:

\begin{equation}
\boldsymbol{\kappa}_{uv}'' = \boldsymbol{\kappa}_{uv}' + \text{SelfAtt}(\boldsymbol{\kappa}_{uv}')
\end{equation}

\noindent This process can be repeated for a number of refinement layers -- in practice we use a maximum of two refinement layers for efficiency. After the attention mechanism, a feed-forward network (FFN) is applied to further transform the representations:

\begin{equation}
\boldsymbol{\kappa}_{uv}^\text{final} = \text{FFN}(\boldsymbol{\kappa}_{uv}'')
\end{equation}

\noindent The same procedure applies when refining relation type representations \( \mathbf{q}_t \) to get \( \mathbf{q}_t^\text{final} \).

The final refined representations \( \boldsymbol{\kappa}_{uv}^\text{final} \) and \( \mathbf{q}_t^\text{final} \) are then used for scoring the relation between the entity pair \( (u, v) \) and the relation type \( t \), as described in Section \ref{sec:scoring-layer}.

\subsection{Scoring Layer}
\label{sec:scoring-layer}

To evaluate whether the relationship between entity pair \( (u, v) \) corresponds to any relation type \( t \) in the set of given relation types \( T \), we calculate the following matching scores:

\begin{equation}
\Phi(u, v) = \{ \phi(u, v, t) \mid t \in T \},
\end{equation}

where

\vspace{-0.4cm}

\begin{equation}
\label{eq:sigmoid-scoring}
\phi(u, v, t) = \sigma(\boldsymbol{\kappa}_{uv}^T \mathbf{q}_t) \in \mathbb{R}
\end{equation}

\noindent In Equation \ref{eq:sigmoid-scoring}, \( \sigma \) denotes a sigmoid activation function. \( \boldsymbol{\kappa}_{uv} \) is the representation of the entity pair representation for entities $u$ and $v$. \( \mathbf{q}_t \) is the relation type representation vector type \( t \). As we train with binary cross-entropy loss, \( \phi(u, v, t) \) can be interpreted as the probability of the entity pair \( (u, v) \) being of type \( t \).

\subsection{Training Dataset Generation}

As discussed in Section \ref{sec:background}, synthetic data generation has been a key enabler for recent improvements in efficient zero-shot NLP models. Due of the difficulty of large-scale manual annotation for relationship classification in particular, synthetic data generation offers a significant improvement in the effectiveness of GLiREL. Our synthetic annotation protocol generates training data for relation classification using an LLM. The goal is to create a flexible relation classification model capable of identifying a broad range of relationship types across various textual domains. It is thus crucial that our training dataset captures diverse relation types.

We follow a methodology similar to \citet{bogdanov2024nunerentityrecognitionencoder}, who utilized the C4 dataset \cite{C4-dataset} to enhance NER. \citet{bogdanov2024nunerentityrecognitionencoder} sampled from C4, an English web crawl dataset widely used for pretraining LLMs, and employed gpt-3.5-turbo to annotate the entity types. Notably, they did not predefine the entity labels, allowing the LLM to extract a diverse set of entities. Similarly, in our work, we employ an LLM to generate a wide variety of relation labels without imposing predefined relation types, thus ensuring a rich and varied set of annotations. To this end, we use a random sample of the Fineweb dataset \cite{penedo2024finewebdatasetsdecantingweb}, another English web crawl dataset chosen for its high quality and diverse material.

We use Mistral 7B-Instruct-v0.3 \cite{jiang2023mistral7b} to annotate every entity pair in every text. The prompt used is shown in the Appendix (Figure \ref{fig:prompt}). Our synthetic dataset contains 63,493 texts with 25,619,624 annotated relations, the majority of which are labeled  \verb|NO RELATION|. For our experiments, we discard those labels that intersect with benchmark labels, in order to strictly maintain the zero-shot paradigm. The dataset is available for public use\footnote{\url{https://huggingface.co/datasets/jackboyla/ZeroRel}}.

\subsection{Extending to Coreference Resolution and Document-Level Relation Classification}


Co-referential reasoning has been demonstrated to significantly enhance the performance of downstream tasks such as extractive question answering, fact verification, and relation extraction \cite{ye-etal-2020-coreferential}. Motivated by this insight, we also investigate GLiREL's performance on document-level relation classification, which requires co-reference resolution to aggregate cluster-level relations, projecting relations to the document level. The results of our experiments are presented in Appendix Section \ref{appendix:coref-and-doc-re}.

\section{Experiments}
\label{sec:experiments}

\subsection{Relation Classification Datasets}

We evaluate GLiREL using the Wiki-ZSL and FewRel benchmarks. \citet{chen2021zsbertzeroshotrelationextraction} derived Wiki-ZSL as a subset of Wiki-KB \cite{sorokin-gurevych-2017-context}, generated through distant supervision. Entities are extracted from complete Wikipedia articles and linked to the Wikidata knowledge base to obtain their relations. FewRel \cite{han2018fewrellargescalesupervisedfewshot} was compiled in a similar manner but underwent additional filtering by crowd workers to enhance data quality and class balance. Although originally designed for few-shot learning, FewRel can be used to benchmark zero-shot relation classification provided that the relation labels in the training and testing sets are disjoint.


Corpus statistics of the Wiki-ZSL and FewRel datasets are summarized in Table \ref{tab:dataset-stats}. Our main results can be seen in Table \ref{tab:zs-relation-classification-results}, with an extended table in the appendix (Table \ref{tab:full-results-appendix}).

\subsection{Zero-Shot Relation Classification Settings}

For each dataset, we randomly select $m$ relations as unseen relations ($m = |Y_u|$) and split the data into training and testing sets, ensuring that these $m$ relations do not appear in the training data so that $Y_s \cap Y_u = \emptyset$. We evaluate using macro precision, recall and F1 score. Experiments are repeated five times with different random selections of unseen relations and train-test splits, and the mean metrics are reported. We vary $m$ to examine its impact on performance and to compare against other models.

\paragraph{Training Details}

For each experiment, we train one model from scratch on the given dataset, and another model is trained following pretraining on our synthetically-annotated dataset. We limit the number of relation type labels prepended to each training instance to 25. For instances where there are less than 25 relation labels, we sample distinct negative labels from the training set. Following \citet{Sainz2023GoLLIEAG}, we introduce regularization by shuffling relation type labels and randomly dropping labels for each instance. We ablate this regularization in Section \ref{sec:random-dropping}. Further training details are provided in the Appendix Section \ref{appendix:training-details}.

\paragraph{Baselines}

We include the results from works described in Section \ref{bkg:zs-re}. We also include the results of OpenAi's GPT-4o model (version \verb|gpt-4o-2024-08-06|) as a baseline for LLM performance on the zero shot relation classification task. In our experiments, we use the prompt as shown in Figure \ref{fig:gpt-prompt} to acquire a prediction for each entity pair in each instance.

\subsection{Results}

GLiREL demonstrates impressive capacity for the zero-shot relation classification task, achieving SoTA performance on both Wiki-ZSL and FewRel. Pretraining on the synthetically annotated dataset shows significant improvement over training from scratch. Additionally, GLiREL is the most succesful model in terms of maintaining performance as the number of unseen labels $m$ increases. GLiREL outperforms GPT-4o at every value of $m$ for each dataset. At $m = 15$, GLiREL is marginally better than the current leading model TMC-BERT. It should be noted that both MC-BERT and TMC-BERT require additional data (entity types and descriptions for each relation type label), as well as one forward pass for each entity pair and label, to achieve their result. GLiREL uses only the given relation type labels, and can classify all entity pair relations in a single forward pass.

\section{Analysis}
\label{sec:ablations}

\subsection{Inference Speed}

We compare the inference speeds of GLiREL against some of the highest-performing ZS relation classification models; TMC-BERT \cite{moeller2024zeroshot} and RelationPrompt \cite{chia-etal-2022-relationprompt}. We run inference for each model on both GPU (one Tesla T4) and CPU. We use WikiZSL and FewRel datasets with number of unseen label $m = 10$ and batch size of 32. Each instance in FewRel contains exactly one entity pair, whereas WikiZSL instances have an average of two (up to a maximum 12). Our performance metric is sentences processed per second. The results can be seen in Table \ref{tab:inference-speed}.

At inference time, the best RelationPrompt model generates $N$ synthetic training examples per unseen label. In our experiments, we set number of synthetic examples $N = 25$, although $N = 250$ is recommended. RelationPrompt NG does not use the generator component to create synthetic training examples. This provides a speed up but at the expense of significant performance deterioration.

RelationPrompt consists of a training data generator (GPT2 with 124M parameters) and an extractor (140M parameters). TMC-BERT has 109M parameters. GLiREL has 467M parameters in total.

\paragraph{Result}

GLiREL maximizes performance while maintaining efficiency for FewRel on CPU, showing a relatively small decrease in throughput when presented with more entity pairs in WikiZSL.  This deterioration is far more pronounced for TMC-BERT, which requires a forward pass for every entity pair, for every candidate label. For RelationPrompt, the generation component poses a significant bottleneck. On GPU, the margin becomes much more apparent, with GLiREL processing 20x more sentences than RelationPrompt and TMC-BERT on Wiki-ZSL.

\begin{table}[ht]
\centering
\resizebox{\columnwidth}{!}{%
\begin{tabular}{l|cc|cc}
\hline
\multirow{2}{*}{Model} & \multicolumn{2}{c|}{Wiki-ZSL} & \multicolumn{2}{c}{FewRel} \\
 & $F_1$ & Speed (sent/s) & $F_1$ & Speed (sent/s) \\
 \hline
\multicolumn{5}{l}{\textbf{CPU}} \\
 RelationPrompt NG& 43.80& 16.41& 55.61&9.27\\
\hline
RelationPrompt & 71.5 & 1.96& 80.0 & 1.59\\
TMC-BERT & 81.23& 0.12& 85.68& 1.91\\
GLiREL & 83.67& \textbf{4.63}& 87.60& \textbf{4.93}\\
\hline
\multicolumn{5}{l}{\textbf{GPU}} \\
 RelationPrompt NG& 43.80& 63.1 & 55.61&59.8 \\
\hline
RelationPrompt & 71.5 & 2.06& 80.0 & 1.72\\
TMC-BERT & 81.23& 1.41& 85.68& 27.81\\
GLiREL & 83.67& \textbf{47.60}& 87.60& \textbf{41.07}\\
\end{tabular}%
}
\caption{Under the number of unseen label $m = 10$ on the zero-shot relation classification task, the comparison between RelationPrompt, TMC-BERT, and GLiREL in F1-score and speed. GLiREL speed is shown in \textbf{bold}.}
\label{tab:inference-speed}
\vspace{-0.2cm} 
\end{table}

\subsection{Ablation Study}

\paragraph{Relation Type Random Dropping}
\label{sec:random-dropping}

We employed a strategy of randomly dropping relation labels in order to vary the number of relation labels during training. This approach aims to increase model robustness to different numbers of labels at inference time. 

\paragraph{Result}
By ablating this component, we see that GLiREL benefits from random dropping on Wiki-ZSL but actually deteriorates on FewRel. This may be caused by the higher number of entity pairs and relation labels in Wiki-ZSL demanding greater generalization of the model, while the single entity pair in FewRel instances provides a cleaner signal and random dropping is unhelpful noise.

\begin{figure}[htbp]
  \centering
  \resizebox{\columnwidth}{!}{
    \includegraphics{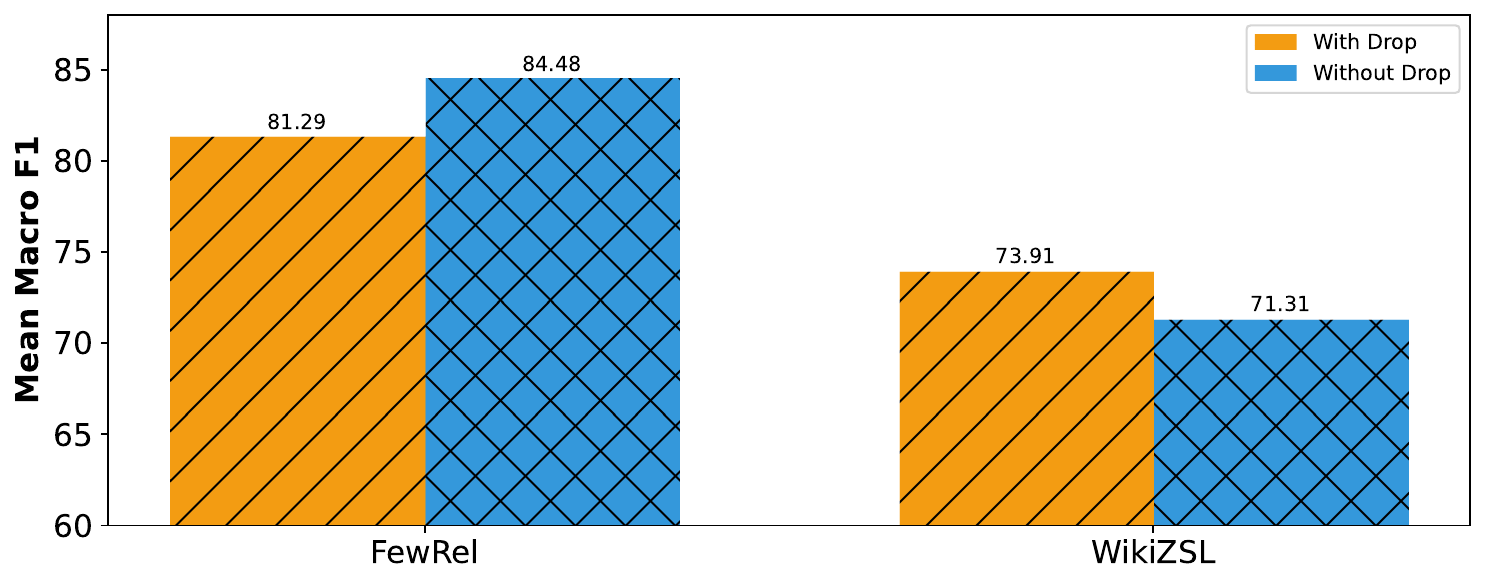}
  }
  \caption{\textbf{Addition of random drop:} The effect of randomly dropping relation labels during training on the FewRel and WikiZSL datasets. Using $m=15$.}
  \label{fig:random-drop-ablation}
\vspace{-0.2cm} 
\end{figure}

\paragraph{Refinement Layers}

The refinement layers as described in Section \ref{sec:refine} aim to enhance the representations of both the entity pair representations and the relation label representations respectively. 

\paragraph{Result}
For the FewRel dataset, we observe benefits from having both prompt and entity pair (relation) refinement layers. Conversely, the model performs best on Wiki-ZSL when no refinement layers are used. As with the random drop ablation, this contrast can be attributed to the difference in entity pairs between the two datasets. The FewRel model benefits from additional depth, as it is modelling only one entity pair interaction per instance. With multiple entity pairs in Wiki-ZSL instances, the cross-attention mechanism may introduce confusing interactions between irrelevant pairs and relations, amplifying signal noise.

\begin{figure}[htp]
  \centering
  \resizebox{\columnwidth}{!}{
    \includegraphics{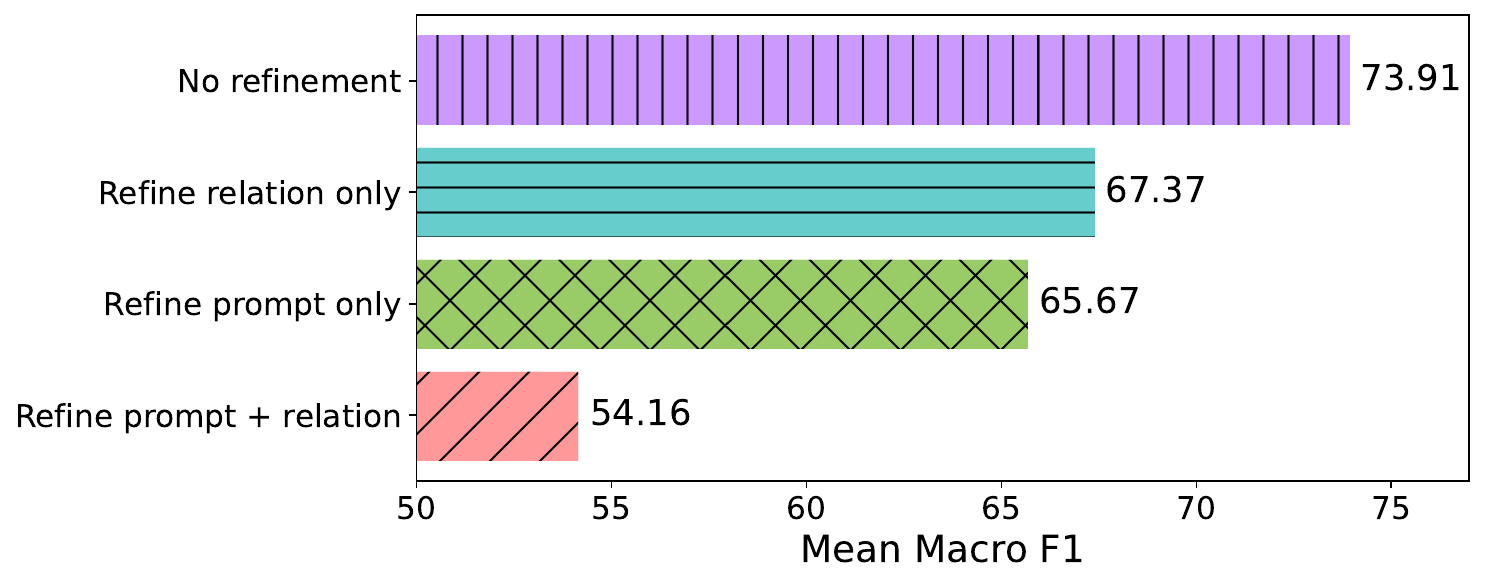}
  }
  \caption{\textbf{Addition of refinement layers:} The effect of adding refine layers for entity pair and relation labels representations. From the WikiZSL dataset, using $m=15$}
  \label{fig:refine-ablation-wiki-zsl}
  \vspace{-0.2cm} 
\end{figure}

\begin{figure}[htp]
\vspace{-0.2cm} 
  \centering
  \resizebox{\columnwidth}{!}{
    \includegraphics{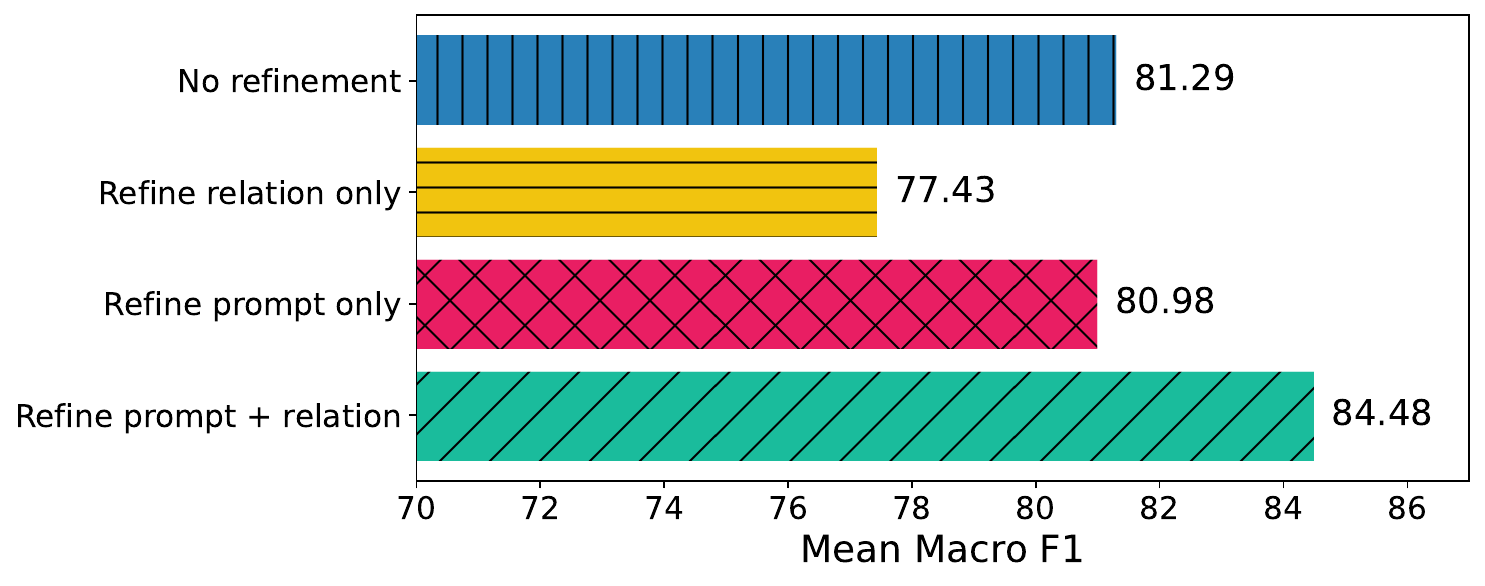}
  }
  \caption{\textbf{Addition of refinement layers:} The effect of adding refine layers for entity pair and relation labels representations. From the FewRel dataset, using $m=15$}
  \label{fig:refine-ablation-few-rel}
  \vspace{-0.2cm} 
\end{figure}

\section{Conclusion}
\label{sec:conclusion}

We have shown that GliREL is a flexible and highly performant approach to zero-shot relation classification (ZSRC), which achieves SoTA results on challenging benchmarks. Unlike other high-performing ZSRC models, GLiREL can classify multiple entity pairs and relation labels in a single input, making it significantly more efficient. Additionally, we have presented a paradigm for generating high-quality, large-scale synthetic datasets for zero-shot relation classification, as well as an effective training protocol. We hope these methods inspire future work in the area of relation classification.

\section{Limitations}
\label{sec:limitations}

As both labels and text are processed in a single forward pass, the number of labels that can be passed in an instance is limited by the model's max sequence length -- in DeBERTa's case this is 512 tokens. It is possible to extend DeBERTa's max positional encoding length to larger values, but that has not been studied here. An avenue for future work that may solve this issue has already been implemented for the GLiNER library, through the use of bi- and poly-encoders.\footnote{\url{https://blog.knowledgator.com/meet-the-new-zero-shot-ner-architecture-30ffc2cb1ee0}} Such architectures enable the use of arbitrary amounts of labels. The embeddings of these labels can be precomputed, which may bring additional efficiency benefits.

The joint encoding of labels and input sequence allows the model to condition label and entity pair representations with respect to one another. This is advantageous in cases where it is important for the model to be aware of all possible labels that can be predicted. However, a limitation of this approach is that the model's performance on one label can be influenced by the order and number of other provided labels.

One benchmark-related issue observed by the authors is that texts in which two entities appear may not provide sufficient evidence for the imputed relationship. For example, an instance in Wiki-ZSL \cite{chen2021zsbertzeroshotrelationextraction} imputes the relation label \verb|P20| (\verb|place of death|) between "Jim Dickinson" and "Memphis" for the following text:
\begin{quote}
"The Pengwins recorded with Rick Derringer at Bearsville Studios in New York and in Memphis with producer Jim Dickinson, and by Columbia and Polygram."
\end{quote}
This is due to the distant annotation of Wiki-ZSL, which uses Wikidata to assign relationships between identified entities. To make a correct prediction, a model would require access to an external knowledge base. This is not implemented in GLiREL or the majority of the methods benchmarked in Table \ref{tab:zs-relation-classification-results}. In light of this, the authors suggest moving away from Wiki-ZSL as one of the primary benchmarks for ZSRC, towards benchmarks that assess a model's ability to extract relationships based solely on the text provided (as seen in FewRel).


\bibliography{custom}

\appendix

\section{Appendix}
\label{sec:appendix}

\subsection{Extended Related Work}
\label{appendix:extended-related-work}

\paragraph{Existing Approaches}
Many systems have addressed relation extraction with varying degrees of success. Earlier works saw CNNs employed in the task of slot filling \cite{adel-etal-2016-comparing-cnn}, a similar task to relation extraction. 

\citet{wang-etal-2022-deepstruct} introduce Deepstruct to improve the structural understanding abilities of language models by pretraining them to generate structures from text on a collection of task-agnostic corpora, enabling zero-shot transfer of knowledge about structure-related tasks. Deepstruct uses this method to achieve state-of-the-art performance on a variety of structured prediction tasks, including RC. 

\citet{riedel-etal-2013-relation-uni-schemas} use matrix factorization and universal schemas to extract relations by leveraging the shared structure between different relations to improve performance.

\citet{rocktaschel-etal-2015-injecting} and \citet{demeester2016liftedruleinjectionrelation} focus on injecting logical background knowledge into embeddings for relation extraction by mapping entity-tuple embeddings into an approximately Boolean space, This method improves generalization and leads to significant performance gains over a matrix factorization baseline.

\citet{zhang-etal-2017-position-aware} have explored the combination of LSTM sequence models with entity position-aware attention to enhance relation extraction. This approach, when coupled with large supervised datasets, has resulted in significant performance improvements in slot-filling tasks.

\paragraph{Question-Answering and Textual Entailment}

\citet{sainz-etal-2021-label} and \citet{obamuyide-vlachos-2018-zero} reformulate relation extraction as an entailment task by using simple verbalizations of relation labels and descriptions. These systems allow for the use of existing textual entailment models and datasets to achieve strong performance in zero-shot and few-shot settings. 

\citet{levy2017zeroshotrelationextractionreading} reduced relation extraction to answering reading comprehension questions by associating natural-language questions with each relation slot. This approach enables the use of neural reading comprehension techniques and supports zero-shot learning by facilitating the extraction of new relation types.

\paragraph{Distant Supervision for Relation Extraction Dataset Construction}

Hand-annotating relation classification (RC) datasets at scale is intractable both because of the size and domain-specificity of relation taxonomies, and especially because of the quadratic number of potential relations in a given text, as a function of the number of named entities in the text.  Foundational research leverages distant supervision to bootstrap training datasets for RC, utilizing open source knowledge bases such as Wikidata \cite{wikidata}, Freebase \cite{freebase} and DBPedia \cite{dbpedia} to obtain high-quality relations between entities, and then mining data sources such as Wikipedia for texts mentioning both head and tail entities to construct training datasets \citep{bunescu_learning_2007,mintz_distant_2009}. 

Distant supervision enables the creation of large scale datasets; however, historical work is still constrained to specific pre-defined label sets, and training data is noisy because inputs are not specifically annotated for particular relations.

\paragraph{Real-world Evaluation of Relation Extraction Models}

\citet{sabo2021revisitingfewshotrelationclassification} critique existing few-shot learning (FSL) datasets for RC, highlighting their unrealistic data distributions, and propose a novel method to create more realistic few-shot test data using the TACRED dataset \cite{zhang-etal-2017-position-aware}, resulting in a new benchmark. Furthermore, they analyze classification schemes in embedding-based nearest-neighbor FSL approaches, proposing a novel scheme that treats the "none-of-the-above" (NOTA) category as learned vectors, improving performance. 

\citet{gao-etal-2019-fewrel-2} present FewRel 2.0 by adding a new, dissimilar domain test set and a NOTA option to the existing FewRel \cite{han2018fewrellargescalesupervisedfewshot} dataset. The authors' experiments reveal that current state-of-the-art models and techniques struggle with these additional challenges that more accurately mirror real-world application of relation extraction models.

\subsection{Tokenization Details}
\label{appendix:tokenization}

The special \verb|[REL]| and \verb|[SEP]| tokens are added as special tokens to the encoder's tokenizer vocabulary. The input sequence from Figure \ref{fig:example-input} is passed to the tokenizer, which joins all elements by whitespace before creating the appropriate encoder-specific subword tokens and input IDs. For example, the label \verb|"participation in"| would be tokenized into the subword tokens: \verb|"particip"|, \verb|"##ation"| and \verb|"## in"|. A mapping from the original input elements in Figure \ref{fig:example-input} to the input IDs is maintained in order to perform subword token pooling of the encoder output. We follow \citet{zaratiana-etal-2022-named}, and perform pooling by taking the vector representation of the first subword token. In the above example, this would correspond to the vector representation for \verb|"particip"|.

With the treatment of one or more tokens as elements, relation type labels can be text of any length. Because the subword tokens of each label are subject to the aforementioned pooling operation, we denote each label using a single index $t_{m}$.

\subsection{GPT-4o Baseline Prompt}

\begin{figure}[htp]
  \centering
  \resizebox{\columnwidth}{!}{
    \includegraphics{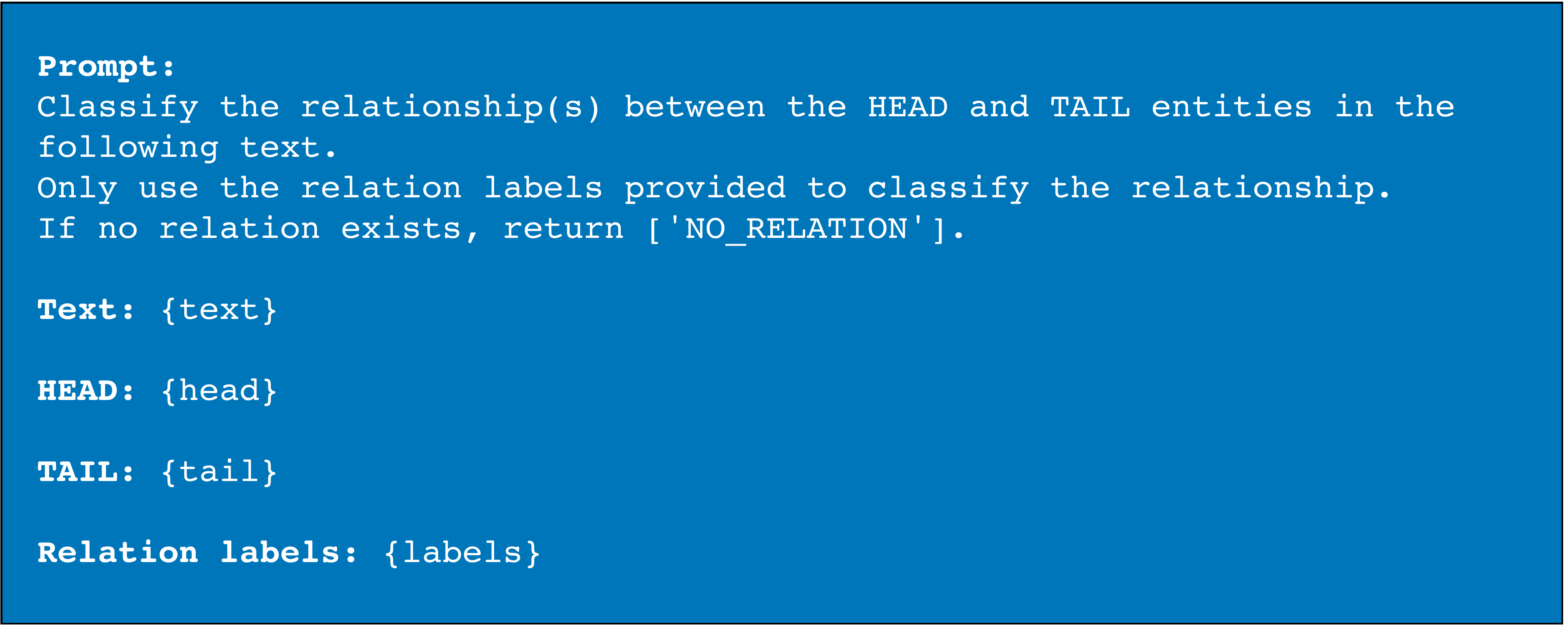}
  }
  \caption{Prompt for synthetic dataset generation.}
  \label{fig:gpt-prompt}
\end{figure}

\subsection{Synthetic Data Generation Details}
\label{appendix:synth-data-gen}

\begin{figure}[htp]
  \centering
  \resizebox{\columnwidth}{!}{
    \includegraphics{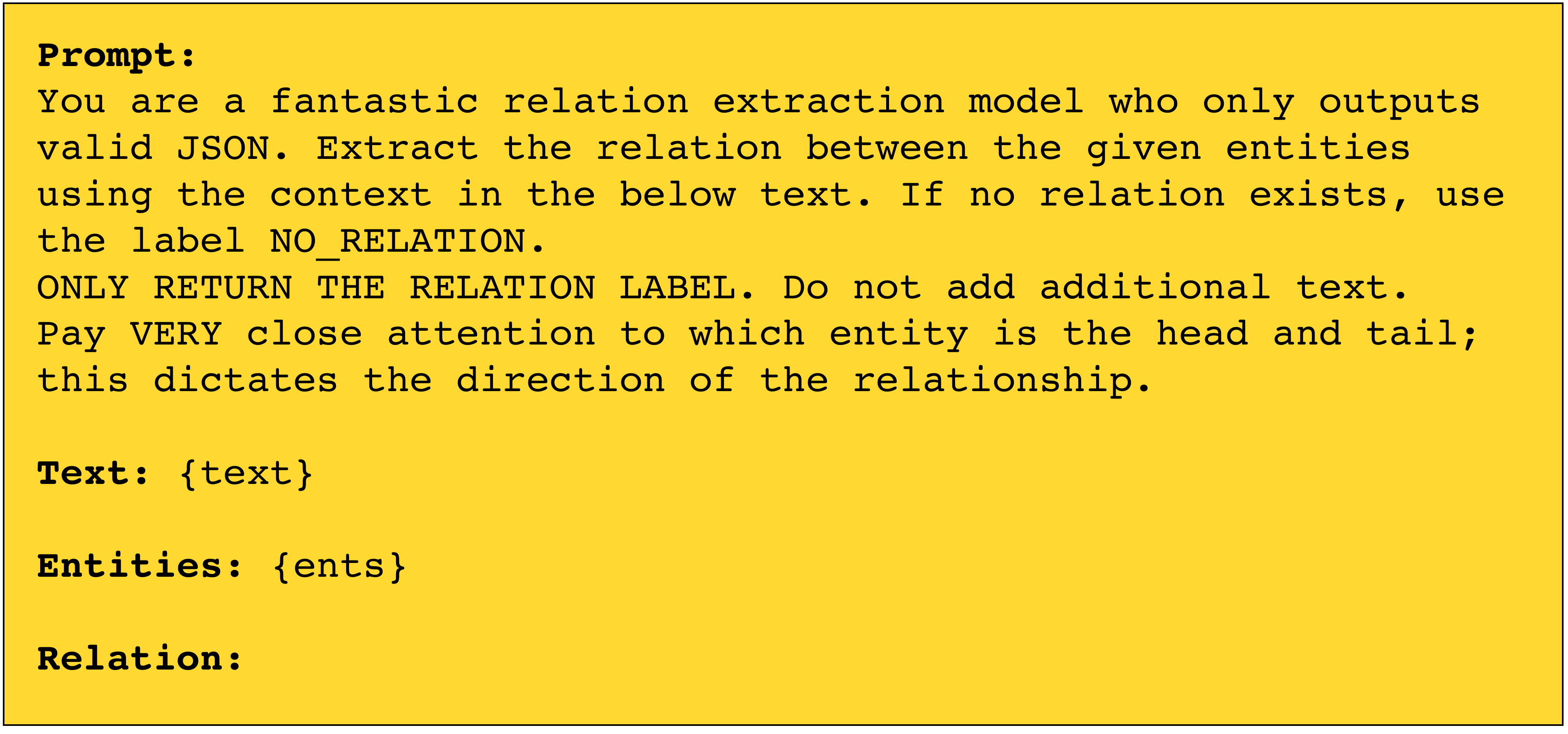}
  }
  \caption{Prompt for synthetic dataset generation.}
  \label{fig:prompt}
\end{figure}

\subsection{Training Setup Details}
\label{appendix:training-details}

For the hyperparameters and configuration of our model, refer to Table \ref{tab:training_setup}. We used the AdamW optimizer \cite{Loshchilov2017DecoupledWD-adamw} with an initial learning rate of $1 \times 10^{-5}$ for the pretrained encoder parameters and $1 \times 10^{-4}$ for the remaining parameters involved in span representation, relation representation and scoring layers. A warmup ratio of 10\% was used with a cosine scheduler. The hidden layer size for all non-encoder layers was set to 768. The batch size was 8, with total number of steps set to 20,000. All experiments were carried out using one NVIDIA Tesla T4 GPU.

\begin{table*}[htbp]
\centering
\caption{Training Setup}
\begin{tabular}{|l|l|}
\hline
\textbf{Hyperparameter/Configuration} & \textbf{Value} \\ \hline
Optimizer                             & AdamW \cite{Loshchilov2017DecoupledWD-adamw} \\ \hline
Initial Learning Rate (Encoder)        & $1 \times 10^{-5}$ \\ \hline
Initial Learning Rate (Other Parameters) & $1 \times 10^{-4}$ \\ \hline
Warmup Ratio                          & 10\% \\ \hline
Scheduler                             & Cosine \\ \hline
Hidden Layer Size (Non-encoder Layers) & 768 \\ \hline
Batch Size                            & 8 \\ \hline
Total Training Steps                  & 20,000 \\ \hline
GPU                                   & NVIDIA Tesla T4 \\ \hline
\end{tabular}
\label{tab:training_setup}
\end{table*}

\subsection{Full Zero-Shot Relation Classification Results}

\begin{table*}[h!]
\centering
\begin{tabular}{c|l|ccc|ccc}
\toprule
\multirow{2}{*}{$m$} & \multirow{2}{*}{Model} & \multicolumn{3}{c|}{Wiki-ZSL} & \multicolumn{3}{c}{FewRel} \\
                     &                        & P     & R     & F1    & P     & R     & F1    \\
\midrule
& CIM \cite{Rocktschel2015ReasoningAE-CIM}                   & 49.63 & 48.81 & 49.22 & 58.05 & 61.92 & 59.92 \\
                     & ZS-BERT \cite{chen2021zsbertzeroshotrelationextraction}                 & 71.54 & 72.39 & 71.96 & 76.96 & 78.86 & 77.90 \\
                     & MICRE w/Llama \cite{li2024metaincontextlearningmakes}                 & 76.46 & 78.53 & 77.48 & 89.34 & 91.88 & 90.59 \\
                     & \citet{tran-etal-2022-improving}     & 87.48 & 77.50 & 82.19 & 87.11 & 86.29 & 86.69 \\
                     & RelationPrompt NG \cite{chia-etal-2022-relationprompt}      & 51.78 & 46.76 & 48.93 & 72.36 & 58.61 & 64.57 \\
                     5& RelationPrompt \cite{chia-etal-2022-relationprompt}         & 70.66 & 83.75 & 76.63 & 90.15 & 88.50 & 89.30 \\
                     & RE-Matching \cite{zhao-etal-2023-rematching}           & 78.19 & 78.41 & 78.30 & 92.82 & 92.34 & 92.58 \\
                     & DSP-ZRSC \cite{lv-etal-2023-dsp}               & 94.1  & 77.1  & 84.8  & 93.4  & 92.5  & 92.9  \\
                     & ZSRE \cite{tran2023enhancing}    & 94.50 & 96.48 & 95.46 & 96.36 & 96.68 & 96.51 \\
                     & MC-BERT \cite{MC-BERT-Lan-et-al}                 & 80.28 & 84.03 & 82.11 & 90.82 & 91.30 & 90.47 \\
                     & TMC-BERT  \cite{moeller2024zeroshot}             & 90.11 & 87.89 & 88.92 & 93.94 & 93.30 & 93.62 \\
 & GPT-4o& 91.24& 72.07& 80.03& 96.75& 83.05&89.20\\
 & GLiREL& 69.88& 65.82& 62.80& 94.56& 89.17&81.21\\
 & GLiREL (+ synthetic pretraining)& 89.41& 80.67& 83.28& 96.84& 93.41&94.20\\
\midrule
& CIM \cite{Rocktschel2015ReasoningAE-CIM}                   & 46.54 & 47.90 & 45.57 & 47.39 & 49.11 & 48.23 \\
                     & ZS-BERT \cite{chen2021zsbertzeroshotrelationextraction}             & 60.51 & 60.98 & 60.74 & 56.92 & 57.59 & 57.25 \\
                     & MICRE w/Llama \cite{li2024metaincontextlearningmakes}                 & 72.36 & 74.88 & 73.60 & 80.67 & 82.31 & 81.48 \\
                     & \citet{tran-etal-2022-improving}     & 71.59 & 64.69 & 67.94 & 64.41 & 62.61 & 63.50 \\
                     & RelationPrompt NG \cite{chia-etal-2022-relationprompt}      & 54.87 & 36.52 & 43.80 & 66.47 & 48.28 & 55.61 \\
                     10& RelationPrompt \cite{chia-etal-2022-relationprompt}         & 68.51 & 74.76 & 71.50 & 80.33 & 79.62 & 79.96 \\
                     & RE-Matching \cite{zhao-etal-2023-rematching}            & 74.39 & 73.54 & 73.96 & 83.21 & 82.64 & 82.93 \\
                     & DSP-ZRSC \cite{lv-etal-2023-dsp}               & 80.0  & 74.0  & 76.9  & 80.7  & 88.0  & 84.2  \\
                     & ZSRE \cite{tran2023enhancing}     & 85.43 & 88.14 & 86.74 & 81.13 & 82.24 & 81.68 \\
                     & MC-BERT \cite{MC-BERT-Lan-et-al}                & 72.81 & 73.96 & 73.38 & 86.57 & 85.27 & 85.92 \\
                     & TMC-BERT \cite{moeller2024zeroshot}              & 81.21 & 81.27 & 81.23 & 84.42 & 84.99 & 85.68 \\
 & GPT-4o& 77.62& 66.14& 68.35& 84.07& 58.00&66.20\\
 & GLiREL& 76.45& 71.80& 68.89& 85.40& 78.29&80.14\\
 & GLiREL (+ synthetic pretraining)& 89.87& 81.56& 83.67& 91.09& 87.42&87.60\\
\midrule
& CIM \cite{Rocktschel2015ReasoningAE-CIM}                    & 29.17 & 30.58 & 29.86 & 31.83 & 33.06 & 32.43 \\
                     & ZS-BERT  \cite{chen2021zsbertzeroshotrelationextraction}                & 34.12 & 34.38 & 34.25 & 35.54 & 38.19 & 36.82 \\
                     & MICRE w/Llama \cite{li2024metaincontextlearningmakes}                 & 67.14 & 68.87 & 67.99 & 73.74 & 75.83 & 74.77 \\
                     & \citet{tran-etal-2022-improving}    & 38.37 & 36.05 & 37.17 & 43.96 & 39.11 & 41.36 \\
                     & RelationPrompt NG \cite{chia-etal-2022-relationprompt}      & 54.45 & 29.43 & 37.45 & 66.49 & 40.05 & 49.38 \\
                     15& RelationPrompt \cite{chia-etal-2022-relationprompt}        & 63.69 & 67.93 & 65.74 & 74.33 & 72.51 & 73.40 \\
                     & RE-Matching  \cite{zhao-etal-2023-rematching}           & 67.31 & 67.33 & 67.32 & 73.80 & 73.52 & 73.66 \\
                     & DSP-ZRSC  \cite{lv-etal-2023-dsp}              & 77.5  & 64.4  & 70.4  & 82.9  & 78.1  & 80.4  \\
                     & ZSRE \cite{tran2023enhancing}    & 64.68 & 65.01 & 65.30 & 66.44 & 69.29 & 67.82 \\
                     & MC-BERT  \cite{MC-BERT-Lan-et-al}              & 65.71 & 67.11 & 66.40 & 80.71 & 79.84 & 80.27 \\
                     & TMC-BERT \cite{moeller2024zeroshot}               & 73.62 & 74.07 & 73.77 & 82.11 & 79.93 & 81.00 \\
 & GPT-4o& 81.04& 32.06& 41.57& 84.42& 65.76&70.70\\
 & GLiREL& 66.14& 65.40& 60.91& 75.76& 71.34&70.40\\
 & GLiREL (+ synthetic pretraining)& 79.44& 74.81& 73.91& 88.14& 84.69&84.48\\
 \bottomrule
\end{tabular}
\caption{Full performance comparison of models on Wiki-ZSL and FewRel datasets for various values of unseen relations $m$. All metrics are averaged on the macro (class) level. }
\label{tab:full-results-appendix}
\end{table*}

\subsection{Coreference Resolution and Document-level Relation Classification}
\label{appendix:coref-and-doc-re}

We conceptualize coreference resolution as a specific case of relation classification, where the coreference relation between two mentions referring to the same entity is represented by a special \verb|SELF| label.

To add coreference resolution ability to GLiREL, we simply include the \verb|SELF| relation type in the label set during training and inference.

In our experiments (Section \ref{sec:experiments}), we evaluate the performance of this approache using the Re-DocRED dataset \cite{tan2023revisitingdocredaddressing}.

\begin{figure}[htbp]
  \centering
  \resizebox{\columnwidth}{!}{
    \includegraphics{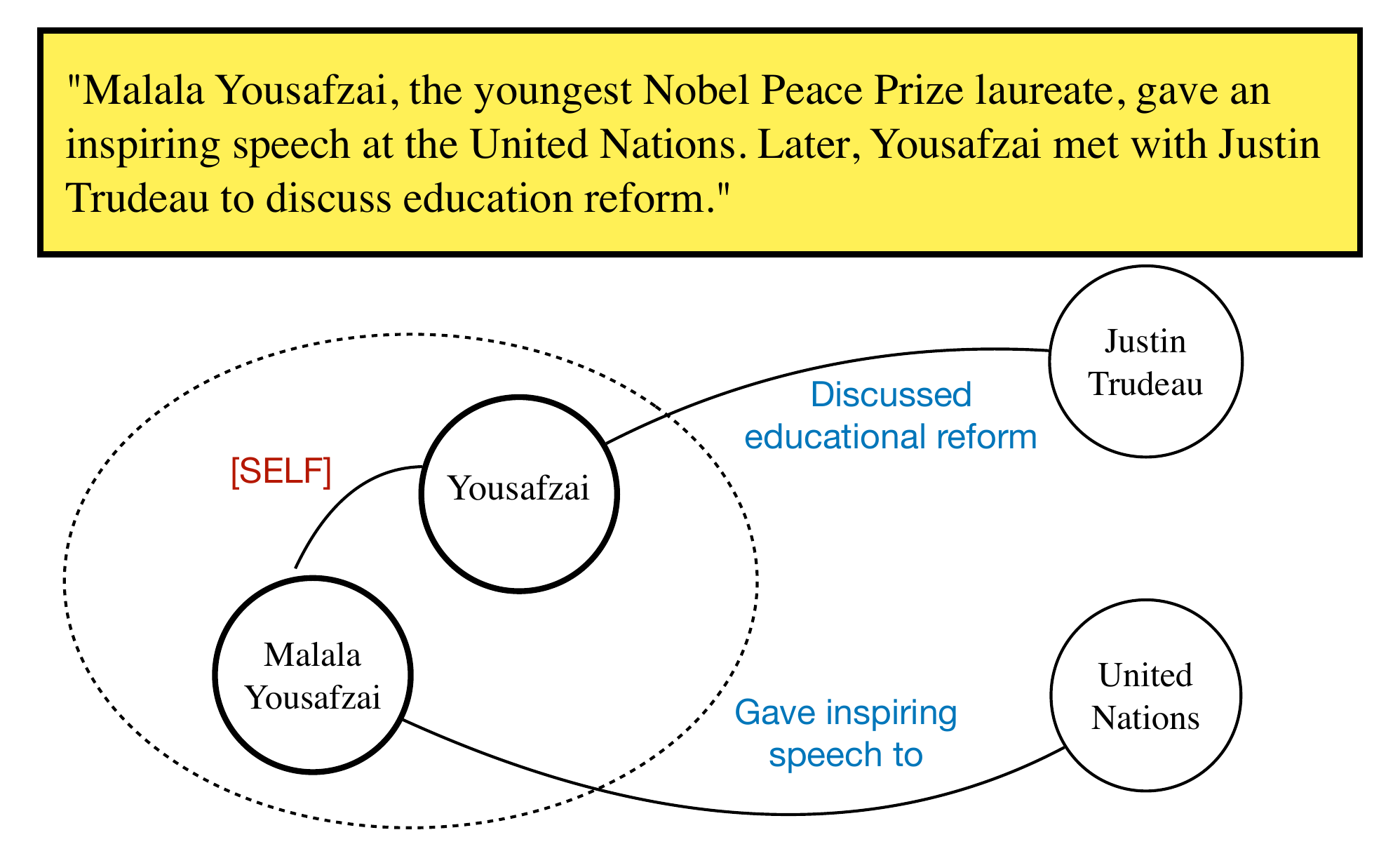}
  }
  \caption{An example of merging entities into clusters and aggregating their relations.}
  \label{fig:coreference-aggregation}
  \vspace{-0.5cm} 
\end{figure}

\paragraph{Document-Level Relation Classification}
When coreference information is available, document-level relation extraction (DocRE) can be achieved by propagating local relations across coreference clusters.  To implement this, we employ a post-processing step that clusters mentions based on the \verb|SELF| relation, akin to the connected components algorithm. The outgoing (non-\verb|SELF|) edges from each mention within a cluster are then interpreted as document-level relations between the resolved entity cluster and other entity clusters in the text. Figure \ref{fig:coreference-aggregation} provides an illustration of this concept.

\paragraph{Number of entity pairs bottleneck}
One bottleneck of the initial GLiREL archictecture is the fact that the number of entity pairs in an instance scales almost quadratically with the number of entities ($N^2 - N$, excluding self-pairs). This becomes a significant memory issue when extending GLiREL to document-level relation classification. To alleviate this issue, we can incorporate a naive windowing method, which only pairs entity within a configured token distance window. With the addition of coreference clusters, the relations predicted within each window can then be aggregated across the document.

We further assess the model's effectiveness on the established DocRE dataset Re-DocRED \cite{tan2023revisitingdocredaddressing}.

To assess GLiREL's performance on the Document-level relation extraction task, we use the Re-DocRED dataset. In one experiment, the model is trained only to predict relations between entities with no coreference \verb|SELF| label. The given (gold) coreference clusters are used to aggregate relations across each document. This is the typical setting for the Re-DocRED benchmark. Additionally, we investigate GLiREL's ability to perform coreference resolution by using the prediction of \verb|SELF| relations between entities to perform coreference.

\paragraph{Baselines}

We compare GLiREL to the strongest models on this benchmark. KD-RoBERTa \cite{tan-etal-2022-document} achieves SoTA results using a RoBERTa-based model, with the addition of an axial attention module to capture interdependencies among entity pairs, and a knowledge distillation framework to make use of large-scale distantly supervised data. \citet{ma-etal-2023-dreeam} perform strongly on Re-DocRED by introducing DREEAM, a method that integrates evidence retrieval (ER) to help the model focus on relevant parts of the document. We also compare LLM-based models -- LMRC \cite{li2024metaincontextlearningmakes}, AutoRE \cite{xue2024autoredocumentlevelrelationextraction}, GenRDK and CoR \cite{sun2024consistencyguidedknowledgeretrieval} -- which were introduced in the Background (Section \ref{sec:background}).

\paragraph{Results}

The results of these approaches are shown in Table \ref{tab:re-docred}. GLiREL achieves competitive performance against finetuned LLMs with over x15 more parameters. However, GLiREL is surpassed by the more specialised framework LMRC, while both BERT-based methods KD-RoBERTa and DREEAM remain significantly better at this benchmark.

Relying on predicted \verb|SELF| relations to perform coreference upon proves to be unreliable, showing poor performance on the benchmark without annotated coreference clusters. 


\begin{table*}
\centering
\begin{tabularx}{0.8\textwidth}{Xcc}
\hline
Method & Ign $F_1$ & $F_1$ \\
\hline
\multicolumn{3}{l}{\textbf{BERT-based}} \\
KD-RoBERTa$_{large}$ \cite{tan-etal-2022-document} & 77.60 & 78.28 \\
DREEAM \cite{ma-etal-2023-dreeam} & \textbf{79.66} & \textbf{80.73} \\
\hline
\multicolumn{3}{l}{\textbf{LLM-based}} \\
CoR \cite{sun2024consistencyguidedknowledgeretrieval} & - & $37.1 \pm 9.2$ \\
GenRDK \cite{sun2024consistencyguidedknowledgeretrieval} & - & $41.3 \pm 8.9$ \\
AutoRE \cite{xue2024autoredocumentlevelrelationextraction} & - & 51.91 \\
LoRA FT LLaMA2-7B-Chat \cite{li2024llmrelationclassifierdocumentlevel} & 52.74 & 53.02 \\
LoRA FT LLaMA2-13B-Chat \cite{li2024llmrelationclassifierdocumentlevel} & 52.15 & 52.45 \\
LMRC-LLaMA2-7B-Chat \cite{li2024llmrelationclassifierdocumentlevel} & 72.33 & 72.92 \\
LMRC-LLaMA2-13B-Chat \cite{li2024llmrelationclassifierdocumentlevel} & 74.08 & 74.63 \\
\hline
\multicolumn{3}{l}{\textbf{GLiREL}} \\
GLiREL (+ gold coref clusters) & 53.24 & 54.13 \\
GLiREL (+ predicted coref clusters) & 25.97 & 25.08 \\
\hline
\end{tabularx}
\caption{Results on the test set of Re-DocRED. Best metrics are shown in \textbf{bold}}
\label{tab:re-docred}
\end{table*}

\end{document}